# Real-time Streaming Perception System for Autonomous Driving

Yongxiang Gu*
*Chengdu Institute of Computer Applications*
*Chinese Academy of Sciences*
Chengdu, Sichuan, China
guyongxiang19@mails.ucas.ac.cn

Qianlei Wang
*Key Laboratory of Flight Techniques and Flight Safety*
*Civil Aviation Flight University of China*
Guanghan, Sichuan, China
qianleiwang@cafuc.edu.cn

Xiaolin Qin*
*Chengdu Institute of Computer Applications*
*Chinese Academy of Sciences*
Chengdu, Sichuan, China
qinxl2001@126.com

*Abstract*—Nowadays, plenty of deep learning technologies are being applied to all aspects of autonomous driving with promising results. Among them, object detection is the key to improve the ability of an autonomous agent to perceive its environment so that it can (re)act. However, previous vision-based object detectors cannot achieve satisfactory performance under real-time driving scenarios. To remedy this, we present the real-time steaming perception system in this paper, which is also the 2nd Place solution of Streaming Perception Challenge (Workshop on Autonomous Driving at CVPR 2021) for the detection-only track. Unlike traditional object detection challenges, which focus mainly on the absolute performance, streaming perception task requires achieving a balance of accuracy and latency, which is crucial for real-time autonomous driving. We adopt YOLOv5 as our basic framework, data augmentation, Bag-of-Freebies, and Transformer are adopted to improve streaming object detection performance with negligible extra inference cost. On the Argoverse-HD test set, our method achieves 33.2 streaming AP (34.6 streaming AP verified by the organizer) under the required hardware. Its performance significantly surpasses the fixed baseline of 13.6 (host team), demonstrating the potentiality of application.

## I. Introduction

Autonomous driving is a comprehensive task that consists of multiple perception level subtasks [1]. In the process of autonomous driving, detecting all potential dangerous targets of the complex surrounding traffic scene in a fast and accurate manner is the basis of safe driving. Meanwhile, detecting traffic signs and road environment is the premise of route planning. Therefore, object detection is a long-term research hotspot in autonomous driving [2].

Thanks to the development of Deep Neural Networks (DNNs), modern object detection approaches have been fundamentally evolved. Currently, object detection algorithms are widely used in various applications. For traffic sign recognition classification , He *et al.* [8] proposed a lightweight convolutional network consisting of simple convolution and pooling layers, achieving high accuracy. Lv *et al.* [9] combined (feature pyramid network) FPN [24] with SSD for improving the small object detection performance of remote sensing images. Wu *et al.* [5] designed a unified framework composed of transfer learning, segmentation, and active classification for vehicle detection. Meanwhile, the emerging research of object detection is developing rapidly. Liu *et al.* [10] introduced Transformer into backbone and reduced model complexity by limiting the receptive field of multi-head attention operations to the fixed, achieving the state-of-the-art performance of 60.6 AP on COCO test-dev [11].

Nevertheless, most approaches merely focus on pushing the standard offline evaluation performance. While for practical application scenarios, accuracy and latency need to be considered comprehensively. To attract researchers to pay more attention on this issue, Streaming Perception Challenge is proposed to score latency and accuracy through streaming AP [7]. As briefly shown in Fig. 1, the real-time processing ability of detection algorithms for 30 FPS continuous video stream is tested with streaming AP. Specifically, participants are required to develop an efficient object detector on Argoverse-HD [15] dataset to achieve higher streaming AP. Since such metric is hardware-dependent, the final score is tested under the official required hardware, i.e., Tesla V100.

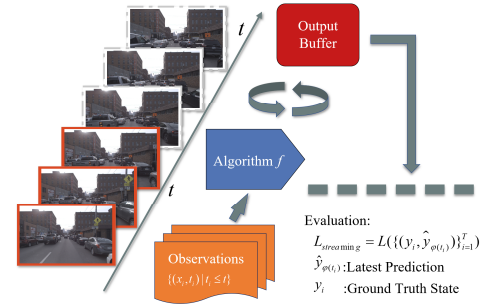

Fig. 1. Streaming perception evaluation

We highlight our principal contributions as follows:

1) Based on the YOLOv5 model, we used data augmentation, Bag-of-Freebies, and Transformer for improving the streaming accuracy with negligible extra inference cost. Furthermore, we get a real-time streaming perception system.

2) Experiments on the Argoverse-HD dataset demonstrate the effectiveness of our method, which achieved 33.2 streaming AP (34.6 streaming AP verified by the organizer) and won the 2nd Place in Streaming Perception Challenge.

## II. Related Work

### A. Modern Object Detection

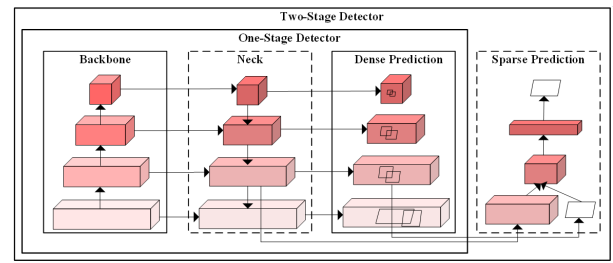

Fig. 2. Structure of modern object detectors

As shown in Fig. 2, a modern detector is usually composed of the following parts: a backbone which outputs the feature map of the whole picture, a neck which fuses feature maps with different scales to obtain multi-scale information, and a head which is used to predict classes and bounding boxes of objects based on the proposals [13]. In general, modern object detection framework can be mainly divided into two categories: one-stage and two-stage. The major difference between two frameworks in structure is that the one-stage generates proposals directly on the feature map, while the two-stage uses RPN to assist in generating proposals.

For two-stage detectors, Faster RCNN [12] is classical and has become the de facto standard architecture, which integrates feature extraction, proposal extraction, bounding box regression, and classification into one network, greatly improving the comprehensive performance. However, the extra sub-network of region proposals (RPN) requires great computational resources, thus limiting the processing speed. By contrast, YOLO is a family of one-stage detectors that can provide good and real-time performance. Therefore, it is nature to consider using the YOLO algorithm to cope with the steaming perception task.

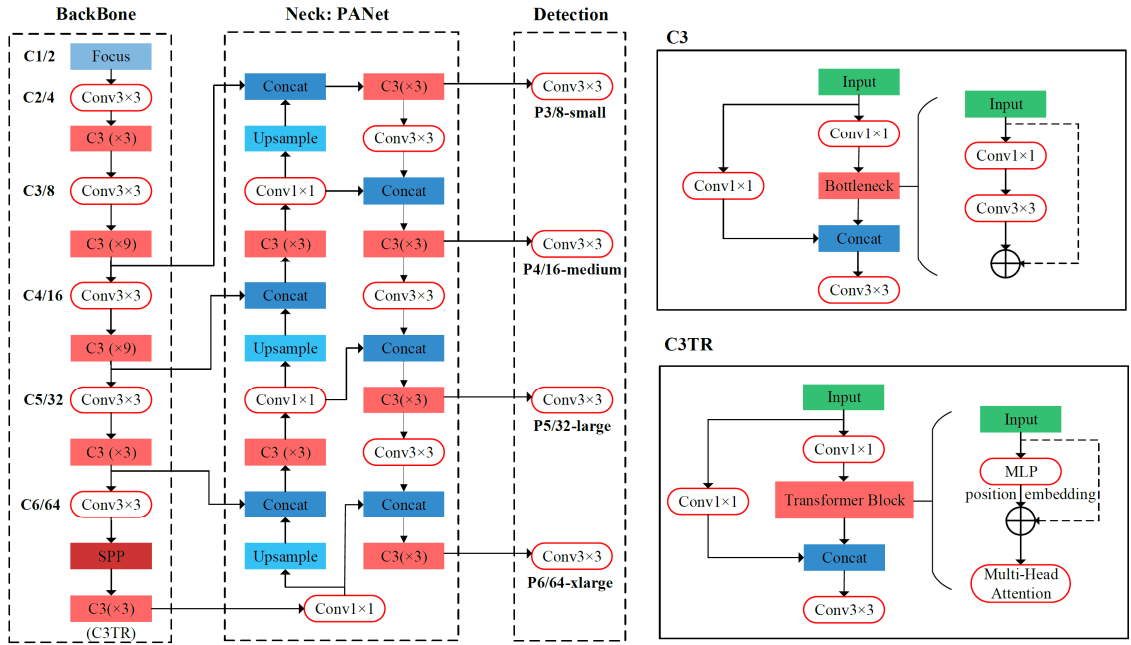

Fig. 3. Overview of YOLOv5m6 architecture

### B. YOLOv5

YOLOv5 [3] is the latest version of YOLO series, which surpasses its predecessors in speed and accuracy. It is proposed by Ultralytics LLC in 2020 and is still in development. By July 2021, the latest release version of YOLOv5 is v5.0, officially called YOLOv5-P6. There are 4 default configurations, i.e., YOLOv5s6 (small), YOLOv5m6 (medium), YOLOv5l6 (large), and YOLOv5x6 (extra), in which parameters increase in sequence. Their network architecture is uniform, while there are differences in network depth and width. In this paper, YOLOv5m6 is chosen to build a baseline for the consideration of model complexity.

As shown in Fig. 3, the overview of YOLOv5m6 can be divided into three parts, i.e., backbone, neck, and detection. In backbone, many effective designs of YOLOv4 and YOLOv3 have been inherited, e.g., spatial pyramid pooling (SPP) [17] module and cross-stage partial (CSP) block [16]. The structure of C3 is consistent with CSP block. By splitting the feature map in calculation, it can significantly reduce computation overhead and optimize the gradient updating process. Specifically, Batch Normalization and SiLU are incorporated into C3 for further enhancing the learning ability. Besides, parameter-less SPP module, which only consists of a series pooling layers with different scales, is adopted to extract multi-scale content information. Therefore, there is little effect on running speed. In neck, PANet [20] is adopted to replace vanilla FPN [21] for fusing features. Comparing with FPN, an extra bottom-up path augmentation (BPA) and the adaptive feature pooling are proposed in PANet for boosting information flow and aggregating features. In detection, its design is consistent with YOLO algorithm, except that the anchor size is calculated through clustering in advance. There are 4 heads in the detection part, which are used for detecting objects with different scales. What's more, YOLOv5 also absorbed other advanced technologies. Among them, Mixup and Mosic [14] are used for data augmentation, GIoU [22] loss is used for enhancing the perception of the instance scale. It is really an excellent work that contributes a lot to the community development.

However, there exist few training tricks provided in YOLOv5, which can also boost performance and own practical significance for engineering projects. To remedy this, we systematically introduce the ignored training techniques in this paper. Besides, we introduce Transformer into YOLOv5 for enhancing the ability of extracting global content information.

### III. METHODOLOGY

Considering the restricted hardware and detection speed requirement, we adopt YOLOv5m6 as our basic model. Bag-of-freebies reported in YOLOv4 [14] impressed us a lot, which can improve the detection performance without increasing the inference cost. Therefore, we mainly focus on techniques in the training stage. Besides, we introduce Transformer in the last block of backbone to enhance the model representation power at a low cost.

## A. Data Augmentation

TABLE I. COMPARISON USING DIFFERENT TRAINING SETS

| Dataset | Training Images | Offline AP |
|---|---|---|
| Argoverse HD | 39384 | 34.0 |
| Agroverse_10per | 3959 | 29.7 |

In the Argoverse-HD dataset, there exist 8 annotated object classes: *person*, *bicycle*, *car*, *motorcycle*, *bus*, *truck*, *traffic light*, and *stop sign*, which are highly related to driving scenarios. Since images are extracted from dozens of videos, it can be found that there exist plenty of approximate frames. We believe that existing training images can provide limited supervision information and models tend to be overfitting when directly trained. To test this conjecture, we extract one frame every ten and build a small dataset, named Agroverse_10per. As shown in Tab. I, we can get close offline results with the same quick research configuration (details will be introduced in the next section) even one-tenth of training images are used.

TABLE II. STATISTICS OF AUGMENTED TRAINING SET

| Dataset | Classes | Training Images |
|---|---|---|
| Agroverse_10per | 8 | 3959 |
| COCO | 8 | 12786 |
| BDD100k | 8 | 70000 |
| Augmented dataset | 8 | 86745 |

To increase sample diversity, we incorporate COCO [11] and BDD100K [6] as external datasets. The former is a common-type dataset and the latter is a dataset of driving scenarios. Since the classes in Argoverse-HD is the subset of COCO, we directly extract the annotated information of 8 classes. However, there is no class named *stop sign* in BDD100K. To solve this problem, we first extract its parent-class *traffic sign* and then predict *stop sign* using previous trained detectors. For other common 7 classes, we directly extract the corresponding annotated information from BDD100k. For the sake of rigor, we manually review the annotated information of *stop sign*. Finally, we get a strong augmented dataset as our final training dataset, whose statistics are shown in TABLE II. Apart from directly augmenting samples with external datasets, we adopt Mosaic [14] and Mixup in the training stage for dynamic data augmentation.

## B. Data Resample

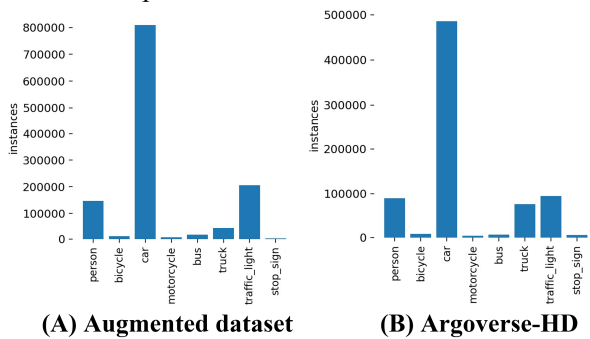

(A) Augmented dataset    (B) Argoverse-HD

Fig. 4. Label Statistics of datasets

As shown in Fig. 4, there exists a class imbalance in the datasets, which is a common issue in objects detection tasks. To remedy this, we design two solutions for data resampling:

(1) We randomly select samples of each class in proportion, and the proportion is inversely proportional to the label number of each class. (2) We use a weighted loss function to make the contribution of each class the same in training. With the same quick research configuration, we got 15.4 and 30.8 AP when using solution one and solution two, respectively. We suspect that the failure of solution one may lie in that the learning rate is set improper when the loss function is multiplied by the weight coefficient.

## C. Structural Re-parameterization

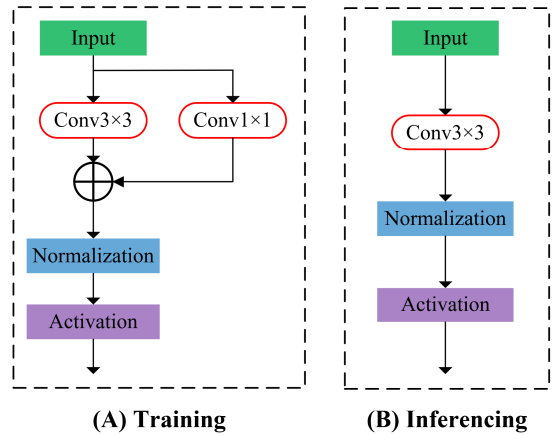

(A) Training    (B) Inferencing

Fig. 5. Diagram of structural re-parameterization

To build a large structure for stronger learning ability as well as larger function space in training, and convert the structure to the small one in inferencing, we use structural re-parameterization [23] in this paper. The mechanism of structural re-parameterization is effective and is easy to implement. Consider there is a basic convolution module without a nonlinear activation function. In training, an identity branch with 1×1 convolution is added to the basic convolution module with 3×3 kernel. While in inferencing, the 1×1 convolution is regarded as a special 3×3 convolution, which is padded with zero. Therefore, two branches can be merged by adding the parameters in their respective positions. What's more, the normalization layer and the activation function can be added normally at the end of this operation. The converting process is shown in Fig. 5.

TABLE III. COMPARISON WITH STRUCTURAL RE-PARAMETERIZATION

| Structural re-parameterization | Params | GFLOPS |
|---|---|---|
| ✓ | 35.4M | 51.3 |
|  | 44.8M | 65.9 |

Specifically, we add an identity branch with $1 \times 1$ convolution to each convolution module to conduct structural re-parameterization and set the initialization value to zero. As shown in TABLE III, under the YOLOv5m6 configuration, we can significantly reduce the number of parameters and calculation with structural re-parameterization in inferencing.

## D. Transformer

Transformer was originally developed in natural language processing (NLP), and has been recently applied to computer vision tasks to achieve the state-of-the-art results [19]. Different from convolution operations, which extract features within a local fixed field, Transformer can adaptively aggregate similar features from a global view using self-attention mechanism. The form of standard self-attention operation can be formulated as follows:

TABLE IV. COMPARISON WITH PROPOSED TECHNOLOGIES TRAINED ON AGROVERSE_10PER

| Data Resample | Structural Re-parameterization | Transformer | Lookahead | Offline AP |
|---|---|---|---|---|
|  |  |  |  | 29.7 |
| ✓ |  |  |  | 30.8 |
|  | ✓ |  |  | 31.7 |
|  |  | ✓ |  | 32.2 |
|  |  |  | ✓ | 30.1 |
| ✓ | ✓ | ✓ | ✓ | **34.5** |

$$Q = f_Q(X), \quad K = f_K(X), \quad V = f_V(X) \quad (1)$$

$$Attention(Q, K, V) = softmax\left(\frac{Q^\top K}{\sqrt{d}}\right)V \quad (2)$$

where $X$ is the input feature map; $f_Q(\cdot)$, $f_K(\cdot)$, $f_V(\cdot)$ are the linear transformation functions; $d$ is the channel dimension; $Q$, $K$, $V$ are abbreviations of query, key and value, respectively.

In this paper, we use the standard Transformer layer in the last block of backbone to obtain global information ( C3TR block in Fig. 3). Since the computational complexity of Transformer is squared with the size of the feature map, our design allows the model to enjoy the benefits of Transformer at a low cost. What's more, other layers of the backbone can reuse the pre-trained weights of the standard YOLOv5m6. Therefore, the efficiency of training can be improved.

*E. Lookahead Optimizer*

Optimization algorithms contribute a lot to the development of deep learning. In 2019, the author of Adam proposed Lookahead optimizer [18], and proved that the performance of Lookahead is better than SGD and Adam through experiments. In Lookahead, fast weight is used to do a quick search and slot weight is used to integrate exploration results. Therefore, the gradient updating process can be more stable and efficient. In this paper, we take Adam as the base optimizer and further build the Lookahead optimizer. Step number and alpha value are set as 5 and 0.5, respectively.

IV. EXPERIMENTS

*A. Dataset and Evaluation*

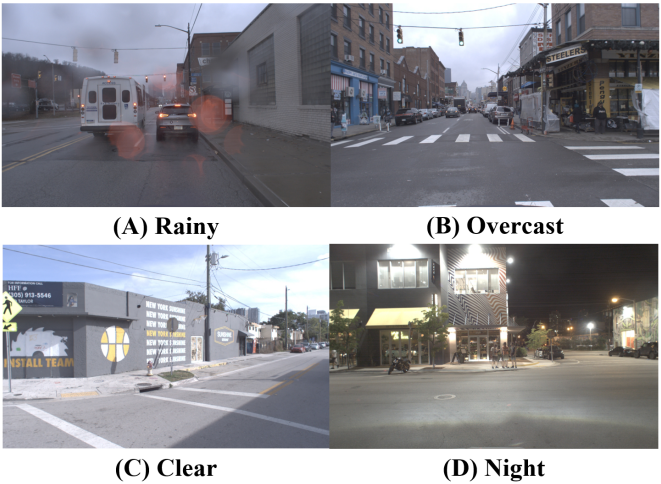

Fig. 6. Examples of Argoverse-HD dataset

(A) Rainy (B) Overcast (C) Clear (D) Night

TABLE V. STATISTICS OF ARGOVERSE-HD DATASET

| Classes | Train Set | Validation Set | Test Set |
|---|---|---|---|
| 8 | 39,384 | 15,062 | 12,507 |

Based upon the autonomous driving dataset Argoverse 1.1 [27], Argoverse-HD dataset is built for streaming evaluation, in which there are a variety of road conditions (Fig. 6). Its image FPS and annotation FPS are both 30, while the image resolution is fixed at 1200×1920. There are totally 250K boxes in Argoverse-HD, therefore, models can get well trained and response detection performance credibly. For doing quick researches, we train models on Agroverse_10per (one-tenth of the train set) to test the proposed technologies. In Streaming Perception Challenge, additional datasets or annotations are allowed to train the model. Therefore, our submitted model is trained on the augmented training set described in the previous section. Since annotations of test set are kept in the organizer, our submitted model is chosen based on the performance of validation set.

For evaluation, there are two different ways, i.e., offline AP and streaming AP. The former is the commonly used matric, which is hardware-independent and is evaluated on validation set, while the latter is hardware-dependent and is used for evaluation in Streaming Perception Challenge [7].

*B. Implementation Details*

In this paper, YOLOv5m6 is chosen as the baseline model. All experiments are conducted on 4 NVIDIA 2080ti GPU and the total batch size is set as 16. In the training stage, Mosaic and Mixup are used for data augmentation. Between them, Mosaic is adopted all time, while the using probability of Mixup is set as 0.24. In the testing stage, we set half precision for acceleration at Tesla V100. When doing quick researches, we only use Agroverse_10per dataset and set image resolution as 720×1280. The number of epochs is set 50. After verifying the effectiveness of all proposed technologies, we use the augmented training set and set image resolution as 1200×1920. To reduce training costs, we first train 100 epochs at 720×1280 resolution and then train 15 epochs at 1920 resolution to ensure the data consistency of training and testing distribution.

*C. Experimental Results*

We first validate each presented technology by doing quick researches. As shown in TABLE IV, the model incorporated with each technology can achieve better performance than the baseline. It should be noted that Transformer yields the highest offline AP improvement, which shows that the global information is crucial for object detection. As expected, the model incorporated with all technologies can achieve the best performance.

TABLE VI.    COMPARSION WITH OTHER METHODS

| Method | Backbone | Image Resolution | Offline AP | sAP | $sAP_{50}$ | $sAP_{75}$ | $sAP_S$ | $sAP_M$ | $sAP_L$ |
|---|---|---|---|---|---|---|---|---|---|
| Mask R-CNN [26] | ResNet50 | 1200×1920 | - | 13.6 | 28.9 | 12.0 | 3.8 | 14.5 | 14.9 |
| Mask R-CNN+RL [25] | Swin-T [10] | - | - | 18.0 | 36.5 | 16.6 | 4.6 | 19.1 | 20.0 |
| YOLOv5 | YOLOv5m6 | 1200×1920 | 34.0 | 21.0 | 35.7 | 20.8 | 4.8 | 20.5 | 30.1 |
| Ours | YOLOv5m6+Transformer | 1200×1920 | **47.6** | **33.2** | **58.6** | **30.9** | **13.3** | **31.9** | **40.0** |

*Note*: sAP means streaming AP, sAP is evaluated on test set while offline AP is evaluated on validation set.

Since our improved method increases negligible extra inference cost, streaming AP improvement should be consistent with offline AP improvement. After doing quick researches, we apply data augmentation, data resample, structural re-parameterization, Transformer, and lookahead optimizer into YOLOv5m6 model. Therefore, we get a real-time streaming perception system. It should be noted that streaming AP depends not only on the hardware but also depend on the input size. The appropriate input size can help to achieve a balance between accuracy and latency. In the proposed method, we set the image resolution as the origin, i.e. 1200×1920. Under the official required hardware, our method achieves 36.1 streaming AP on validation set and 33.2 streaming AP (34.6 streaming AP verified by the organizer) on test set. As shown in TABLE VI, our method outperforms other competitive detectors by a large margin, demonstrating the potentiality of application for real-time autonomous driving.

## V. CONCLUSION

In this paper, we propose the real-time streaming perception system for autonomous driving, which is also the 2nd Place solution of Streaming Perception Challenge for the detection-only track. To achieve a balance between accuracy and latency, we adopt YOLOv5 as our basic framework, data augmentation, Bag-of-Freebies, and Transformer are adopted to improve the streaming detection performance with negligible extra inference cost. Effective training tricks can boost performance and own practical significance for engineering projects. Therefore, we introduce them in detail and make the brief analysis. Besides, we introduce Transformer in the last block of backbone to enhance the model ability of extracting global information at a low cost. By doing quick researches, we validate the effectiveness of each proposed technology. What's more, results on Streaming Perception Challenge demonstrate the outstanding performance of our method.

## ACKNOWLEDGMENT

This work is supported by Sichuan Science and Technology Program (2019ZDZX0006, 2020YFG0010), the West Light Foundation of Chinese Academy of Sciences, National Academy of Science Alliance Collaborative Program (Chengdu Branch of Chinese Academy of Sciences - Chongqing Academy of Science and Technology) and Science and Technology Service Network Initiative (KFJ-STS-QYZD-2021-21-001).